\documentclass[conference]{IEEEtran}

\IEEEoverridecommandlockouts
\usepackage{cite}
\usepackage{amsmath,amssymb,amsfonts}
\usepackage{algorithmic}
\usepackage{graphicx}
\usepackage{multirow}
\usepackage{textcomp}
\usepackage{xcolor}
\def\BibTeX{{\rm B\kern-.05em{\sc i\kern-.025em b}\kern-.08em
    T\kern-.1667em\lower.7ex\hbox{E}\kern-.125emX}}
\begin{document}

\title{Dual Information Speech Language Models for Emotional Conversations \\
\thanks{Presented at IEEE ICME 2025\\ Work done when Wenze Xu interned at Mashang Consumer Finance Co., Ltd. \\The Appendices are available in: https://drive.google.com/drive/folders/
1LL1uKQK5nL8IFlu6XpPH3-bvFQtLaABO?usp=sharing}
}

\author{\IEEEauthorblockN{Chun Wang\IEEEauthorrefmark{1}\IEEEauthorrefmark{3},
Chenyang Liu\IEEEauthorrefmark{1},
Wenze Xu\IEEEauthorrefmark{1}\IEEEauthorrefmark{2},  
Weihong Deng\IEEEauthorrefmark{1}\IEEEauthorrefmark{3}
}
\IEEEauthorblockA{\IEEEauthorrefmark{1}Mashang Consumer Finance Co., Ltd., Chongqing, China}
\IEEEauthorblockA{\IEEEauthorrefmark{2}The University of Sydney, Sydney, Australia}
\IEEEauthorblockA{\IEEEauthorrefmark{3}lukewang25@live.cn; weihong.deng@msxf.com} 
}

\maketitle

\begin{abstract}
Conversational systems relying on text-based large language models (LLMs) often overlook paralinguistic cues, essential for understanding emotions and intentions. Speech-language models (SLMs), which use speech as input, are emerging as a promising solution. However, SLMs built by extending frozen LLMs struggle to capture paralinguistic information and exhibit reduced context understanding. We identify entangled information and improper training strategies as key issues. To address these issues, we propose two heterogeneous adapters and suggest a weakly supervised training strategy. Our approach disentangles paralinguistic and linguistic information, enabling SLMs to interpret speech through structured representations. It also preserves contextual understanding by avoiding the generation of task-specific vectors through controlled randomness. This approach trains only the adapters on common datasets, ensuring parameter and data efficiency. Experiments demonstrate competitive performance in emotional conversation tasks, showcasing the model's ability to effectively integrate both paralinguistic and linguistic information within contextual settings.
\end{abstract}

\begin{IEEEkeywords}
conversation, paralinguistic, linguistic, emotion
\end{IEEEkeywords}

\section{Introduction}
\label{sec:intro}
Conversational systems are essential for making human-computer interactions both effective and engaging. To develop an emotional conversational system, understanding conversational context and user expression are key capabilities \cite{liu-etal-2021-towards}. Modern conversational systems often use a text-based large language model (LLM) as their core. By leveraging the LLM's language processing capability, these systems have achieved a long-term understanding of human interactions, leading to coherent responses in multi-turn conversation scenarios.

\begin{figure}[t]
\centering
\begin{minipage}[b]{1.0\linewidth}
  \centering
  \centerline{\includegraphics[width=6.5cm]{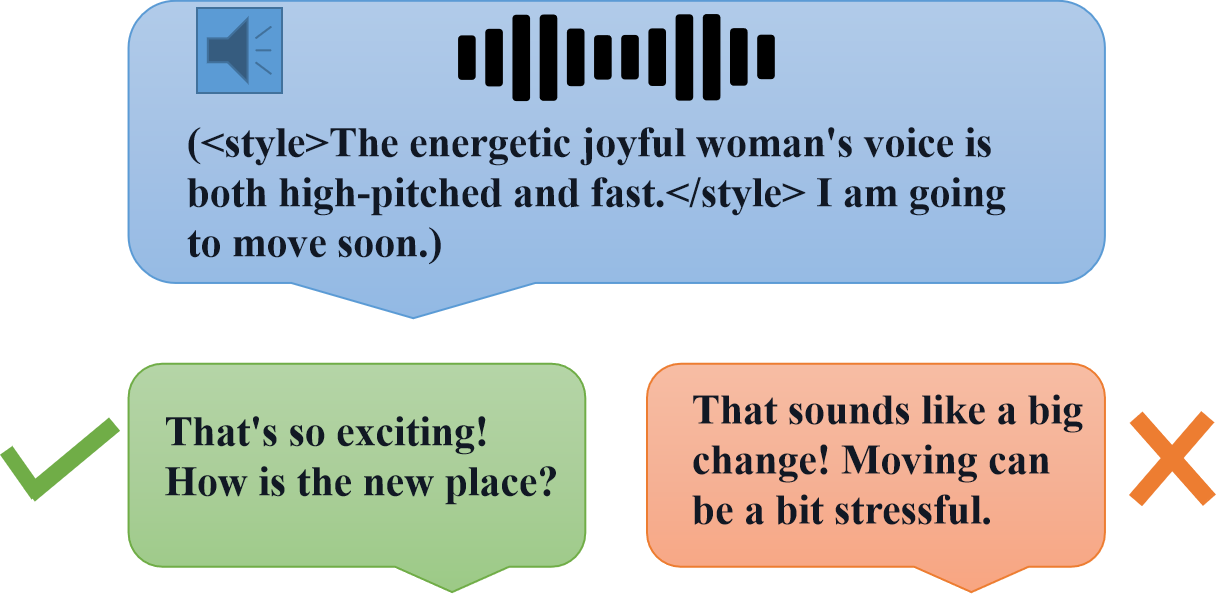}}
\end{minipage}
\caption{An illustrative example of how understanding paralinguistic information influences response generation.}
\label{fig:case}
\vspace{-1em}
\end{figure}

However, text-based LLMs may misinterpret user expressions as they consider only linguistic information \cite{lin-etal-2024-advancing}. Paralinguistic information, such as pitch and speed, is crucial for understanding emotions and intentions \cite{10446933}. As shown in Fig.~\ref{fig:case}, understanding paralinguistic information enables the system to respond appropriately. Otherwise, conversational systems may overlook the user's sentiment, leading to misunderstandings. This underscores the importance of including speech in the input for a comprehensive understanding of users' expressions, ensuring conversational systems generate more accurate responses. Consequently, recent developments have focused on Speech-Language Models (SLMs) \cite{liu-etal-2021-towards}\cite{lin-etal-2024-advancing}\cite{Echat}\cite{BLSPEmo}.

\begin{figure}[tb]
\centering
\begin{minipage}[b]{0.62\linewidth}
  \centering
  \centerline{\includegraphics[width=1.0\linewidth]{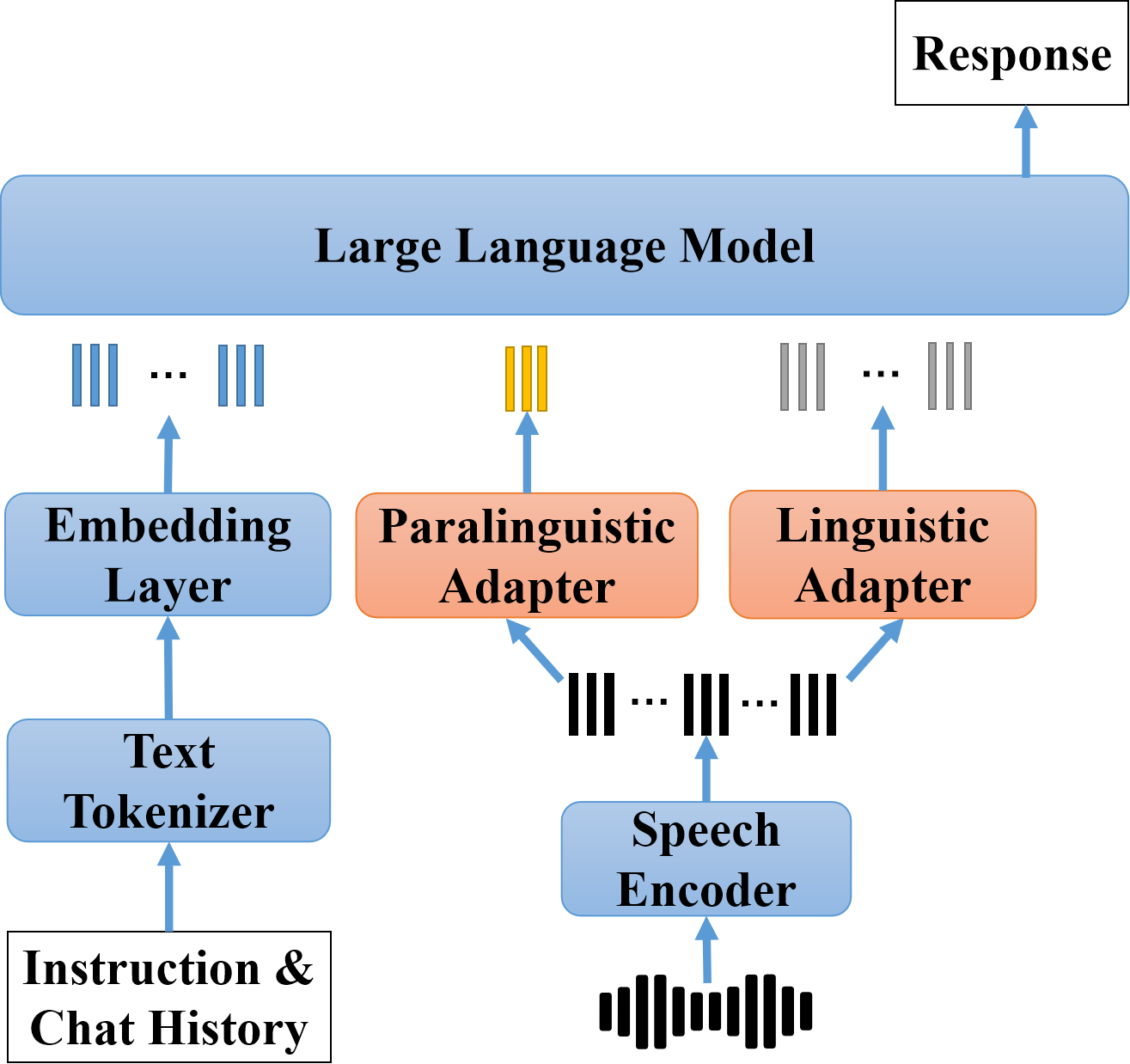}}
\end{minipage}
\caption{Overview of the SLM model architecture. Learnable modules in red, while frozen modules in blue.}
\label{fig:architectureours}
\vspace{-1em}
\end{figure}

However, developing SLMs presents significant challenges. One approach is to build a speech-text foundation model that natively processes and understands spoken language. While effective, this approach requires extensive multimodal data and high computational resources, which limit its feasibility \cite{WavChat}.

A more feasible approach is to augment existing text-based LLMs with speech understanding capabilities. This involves integrating a speech encoder with a text LLM, connecting the two using an adapter module. Previous works using this approach have shown that SLMs built with frozen LLMs can effectively understand linguistic information from speech\cite{10445874}\cite{SLAMASR}. However, they still face two main challenges:
\begin{itemize} 
	\item \textit{Paralinguistic Info Omission}: Trained SLMs often fail to perceive paralinguistic information conveyed in speech. 
	\item \textit{Context Omission}: Trained SLMs often exhibit a reduced understanding of conversational context. 
\end{itemize}

We identify information entanglement and improper training strategies as key issues. Current methods \cite{Echat}\cite{BLSPEmo}\cite{kang2024frozenlargelanguagemodels} often rely on a single adapter to encode both linguistic and paralinguistic information. When these embeddings are projected into the input embedding space of a frozen LLM, this text-trained space naturally prioritizes linguistic content, neglecting paralinguistic aspects. Additionally, existing approaches commonly train SLMs with a limited range of task types in an instruction-tuning manner, without sufficient consideration of the risk of adapters generating embeddings that degrade into task-specific vectors \cite{li-liang-2021-prefix}. This can result in task overfitting and impair the model's contextual understanding.

In this work, we present an efficient approach for building SLMs designed for emotional conversational systems. To address the omission of paralinguistic information, we propose representing paralinguistic and linguistic information in a structured way, enabling the SLM to perceive them separately through distinct mechanisms. This structured representation is achieved by employing a dedicated dual-adapter model architecture and a weakly supervised training strategy.

As shown in Fig.~\ref{fig:architectureours}, we employ two heterogeneous adapters to separately generate paralinguistic and linguistic embeddings. Paralinguistic information, which mostly remains consistent throughout an utterance \cite{lin-etal-2024-advancing}, is captured by fixed-length embeddings generated by the paralinguistic adapter. Conversely, linguistic information, which varies over time, is captured by utterance-length-dependent embeddings generated by the linguistic adapter. The structural heterogeneity ensures that the two adapters prioritize capturing information from distinct perspectives, thereby facilitating easier disentanglement.

To train the adapters for disentangling information, we employ a weakly supervised training strategy. At its core lies Equivalence Replacement Regularization (ERR), designed to ensure that the SLM generates responses based on the relevant embeddings. ERR operates on the principle that for linguistic tasks, which primarily rely on linguistic information, the SLM must perform consistently using only linguistic embeddings, regardless of the presence or source of paralinguistic embeddings. During linguistic adapter training, the paralinguistic adapter remains frozen, and linguistic embeddings are randomly paired with paralinguistic embeddings derived from text, speech, or none. A similar method is applied when training the paralinguistic adapter for paralinguistic tasks.

In this design, the two adapters generate embeddings that convey complementary information, jointly providing a structured representation of speech. These embeddings influence the SLM through distinct mechanisms: linguistic embeddings directly replace the corresponding text content embeddings \cite{audiochatllama}, while paralinguistic embeddings act as soft prompts \cite{PTuningv2}, guiding the SLM's attention toward paralinguistic aspects.

To preserve the model's contextual understanding, we incorporate two forms of randomness to prevent adapters from generating task-specific vectors. Positional randomness is achieved by using multi-turn conversation data with dynamic context lengths during training. As illustrated in Fig.~\ref{fig:architectureours}, context embeddings are positioned at the beginning of the input sequence, preceding the paralinguistic and linguistic embeddings, with their placements significantly varied. This positional randomness, combined with the additional combination randomness introduced through ERR sampling, mitigates overfitting by disrupting occurrence patterns.

Experimental results demonstrate that SLMs trained with the proposed approach effectively perceive both paralinguistic and linguistic information while understanding context. They achieve competitive performance when compared to leading SLMs in emotional conversational scenarios, underscoring the effectiveness of our approach.

The remainder of this paper is organized as follows: Sec. 2 reviews related works. Sec. 3 describes the proposed methods. Sec. 4 and Sec. 5 present the experimental setup and results, respectively. Finally, Sec. 6 concludes the paper.

\section{Related Work}
\label{sec:relatedwork}

\subsection{Speech Large Language Models}
\label{ssec:sllm}
Several works primarily focus on adapting LLMs for specific tasks, such as speech recognition \cite{10445874}\cite{SLAMASR} or speech captioning \cite{SECap}\cite{10445977}. Recently, some research aims to extend LLMs with a general-purpose speech input interface \cite{kang2024frozenlargelanguagemodels}\cite{audiochatllama}\cite{Qwen2Audio}. Our work is similar to these efforts in maintaining LLMs' capabilities, but with a focus on turn-based emotional conversation. Our work is closely related to efforts enabling LLMs to understand paralinguistic information in dialogue. Models like ParalinGPT \cite{10446933} and the Spoken-LLM framework \cite{lin-etal-2024-advancing} augment text content with paralinguistic embeddings from speech, generating appropriate responses. Recent works like E-chat \cite{Echat}, BLSP-Emo \cite{BLSPEmo}, and SpeechEmotionLlama \cite{kang2024frozenlargelanguagemodels} perceive both paralinguistic and linguistic information from speech. However, these models use a single adapter to generate embeddings representing both linguistic and paralinguistic information, necessitating additional training of the speech encoder and/or the LLM to enable the SLM to perceive both types of information, which may unexpectedly alter their inherent capabilities. In our work, we construct the SLM with both the underlying speech encoder and LLM frozen, enabling the SLM to perceive both types of information from speech.

\subsection{Prompt Tuning}
\label{ssec:pt}
Prompt tuning refers to a class of Parameter-Efficient Fine-Tuning (PEFT) methods that integrate trainable continuous vectors into the input, optimized to influence the LLM's response without modifying the entire model. Prefix Tuning \cite{li-liang-2021-prefix} adds trainable continuous vectors at the input's beginning, known as the prefix. Further efforts like Prompt Tuning \cite{lester-etal-2021-power} and P-Tuning \cite{PTuningv2} introduce soft prompts, allowing these vectors to be added within the input sequence, providing nuanced control. Additionally, Multitask Prompt Tuning \cite{wang2023multitask} enhances the model's ability to handle multiple tasks simultaneously by optimizing shared prompts. However, these efforts aim to adapt pre-trained LLMs to specific tasks, causing these vectors to become task-specific \cite{li-liang-2021-prefix}.

In our work, we employ the soft prompt mechanism to enable frozen LLMs to perceive paralinguistic information from paralinguistic embeddings. To avoid side effects, we introduce randomness during training to prevent the two adapters from producing embeddings that degenerate into task-specific vectors, which can lead to task overfitting and reduce the understanding of instructions and context.

\section{Methods}
This section introduces the model architecture, describes the training tasks, and details the weakly supervised training strategy used for developing the SLM.

\subsection{Model Architecture}
As illustrated in Fig.~\ref{fig:architectureours}, the proposed SLM consists of four key components: a speech encoder, a paralinguistic adapter, a linguistic adapter, and an LLM with its tokenizer and embedding layer exposed.

The SLM takes two inputs: a text prompt $\mathbf{X}^{\mathbf{T}}$ and a speech signal $\mathbf{X}^{\mathbf{S}}$. The text embeddings $\mathbf{E}^{\mathbf{T}} \in \mathbb{R}^{n_t \times d_l}$ are generated using the tokenizer and embedding layer of the LLM, formulated as:

\begin{equation} \mathbf{E}^{\mathbf{T}} = \text{EmbedLayer}(\text{Tokenizer}(\mathbf{X}^{\mathbf{T}})), \end{equation}

where $n_t$ represents the text embedding sequence length and $d_l$ is the input embedding dimension of the LLM. Similarly, the speech embeddings $\mathbf{E}^{\mathbf{S}} \in \mathbb{R}^{n_s \times d_s}$ are produced via the speech encoder, formulated as:

\begin{equation} \mathbf{E}^{\mathbf{S}} = \text{SpeechEncoder}(\mathbf{X}^{\mathbf{S}}), \end{equation}

where $n_s$ denotes the speech embedding sequence length and $d_s$ is the dimension of the speech embeddings.

The speech embeddings $\mathbf{E}^{\mathbf{S}}$ are then processed by two adapters: the paralinguistic adapter $\mathbf{A}$ and the linguistic adapter $\mathbf{C}$. This results in paralinguistic embeddings $\mathbf{E}^{\mathbf{A}} \in \mathbb{R}^{n_a \times d_l}$ and linguistic embeddings $\mathbf{E}^{\mathbf{C}} \in \mathbb{R}^{n_c \times d_l}$, where $n_a$ and $n_c$ are the respective sequence lengths.

These embeddings are concatenated along with text embeddings and fed into the LLM to generate the text response $\mathbf{R}^{\mathbf{T}}$, formulated with a slight abuse of notation as:

\begin{equation} \text{ChatTemplate}(\mathbf{E}^{\mathbf{T}}, \mathbf{E}^{\mathbf{A}}, \mathbf{E}^{\mathbf{C}}) \rightarrow \text{LLM} \rightarrow \mathbf{R}^{\mathbf{T}}. \end{equation}

\textbf{Paralinguistic Adapter}. The paralinguistic adapter is designed to capture the paralinguistic information of an utterance, producing paralinguistic embeddings $\mathbf{E}^{\mathbf{A}}$. It processes the speech embeddings $\mathbf{E}^{\mathbf{S}}$ using a compact transformer block, which includes multiple transformer layers with multi-head self-attention and dropout. The processed sequence is then adaptively pooled along the sequence dimension to a fixed length $n_a$. A linear layer subsequently projects the pooled embeddings into the input dimension of the LLM, resulting in a fixed-length sequence that conveys paralinguistic information:

\begin{equation} \mathbf{E}^{\mathbf{A}} = \text{Linear}(\text{Pool}(\text{Transformer}(\mathbf{E}^{\mathbf{S}}))). \end{equation}

\textbf{Linguistic Adapter}. The linguistic adapter is designed to capture the linguistic information of an utterance, producing linguistic embeddings $\mathbf{E}^{\mathbf{C}}$. Given the inherent sparsity of speech embeddings compared to text embeddings, we transform the speech embeddings sequence $\mathbf{E}^{\mathbf{S}}$ into a more compact representation, as outlined in \cite{10445874}\cite{SLAMASR}. Specifically, the speech embeddings sequence $\mathbf{E}^{\mathbf{S}}$ is converted into a compact sequence $\mathbf{H}^{\mathbf{S}} \in \mathbb{R}^{(n_s // k) \times (d_s \cdot k)}$ by concatenating every $k$ adjacent speech embeddings $\mathbf{e}^{\mathbf{S}}_{i}, \mathbf{e}^{\mathbf{S}}_{i+1}, \ldots, \mathbf{e}^{\mathbf{S}}_{i+k-1}$ into a single compact embedding $\mathbf{h}^{\mathbf{S}}_{i}$. This compact sequence $\mathbf{H}^{\mathbf{S}}$ is then processed through two linear layers with an intermediate ReLU activation function and a hidden layer dimension $d_h$, producing the final linguistic embeddings:

\begin{equation} \mathbf{E}^{\mathbf{C}} = \text{Linear}(\text{ReLU}(\text{Linear}(\mathbf{H}^{\mathbf{S}}))). \end{equation}

The architectures of the two adapters are designed with consideration for the nature of the targeted information. Their heterogeneous structures enable the capture of different types of information, making it easier to focus on specific aspects. During training, to preserve model capabilities, the pretrained speech encoder and the LLM are kept frozen, with only the two adapters being learnable.

\subsection{Training Tasks}
This section outlines the training tasks, classifying them into three categories based on the primary information they rely on. Each task type serves a distinct purpose during training.

\textbf{Linguistic Tasks}. Linguistic tasks primarily rely on linguistic information. To enable the understanding of linguistic information, we train the SLM with the Automatic Speech Recognition (ASR) task \cite{SLAMASR}, where the SLM is instructed to produce the transcript of an utterance. 

\textbf{Paralinguistic Tasks}. Paralinguistic tasks primarily rely on paralinguistic information. To enhance the perception of paralinguistic information, we train the SLM using various speech attribute classification tasks \cite{ando24_interspeech}, including gender, pitch, speed, energy, and emotion. Each task prompt consists of an instruction and a list of choices, with the expected response in sentence format. 

\textbf{Dual Information Tasks}. Dual information tasks rely on both paralinguistic and linguistic information. To enable the adaptive utilization of these types of information, we train the SLM using a style-aware behavior alignment task \cite{BLSPEmo}\cite{kang2024frozenlargelanguagemodels}\cite{audiochatllama}. This approach involves aligning the responses of the SLM with responses of the underlying LLM given equivalent inputs, typically the speech input and its styled transcript. 

All three types of tasks are chosen and applied with practical considerations. First, we focus on real tasks and classify them based on the primary information they rely on, without requiring any task to exclusively depend on a single aspect of information, as this is difficult to achieve. Second, tasks are selected and designed with clear, determinable answers. However, open-ended tasks increase the risk of adapters generating task-specific vectors, as evidenced by responses that mimic sentence structures and word choices from the training data \cite{10889444}. This risk is further heightened when annotations significantly differ from the underlying LLM’s natural response style, potentially altering the frozen LLM’s behavior. Third, the natural tendencies of the LLM are carefully considered when designing tasks. For example, instruction-following LLMs typically produce sentence-based responses, and different LLMs adopt distinct instruction templates. Additional details and examples can be found in Sec.~I of the Appendices.

\subsection{Training Strategies}
This section introduces the proposed weakly supervised training strategy, implemented as a three-stage instruction-tuning process.

\textbf{Stage 1}: This stage focuses on enabling the SLM to generate responses based on speech input. The SLM is jointly trained on both linguistic and paralinguistic tasks, with its two adapters being learnable. While the SLMs trained after this stage are capable of generating responses, they show limited context understanding, and the information remains entangled.

\begin{figure}[tb]
\centering
\begin{minipage}[b]{1.0\linewidth}
  \centering
  \centerline{\includegraphics[width=0.8\linewidth]{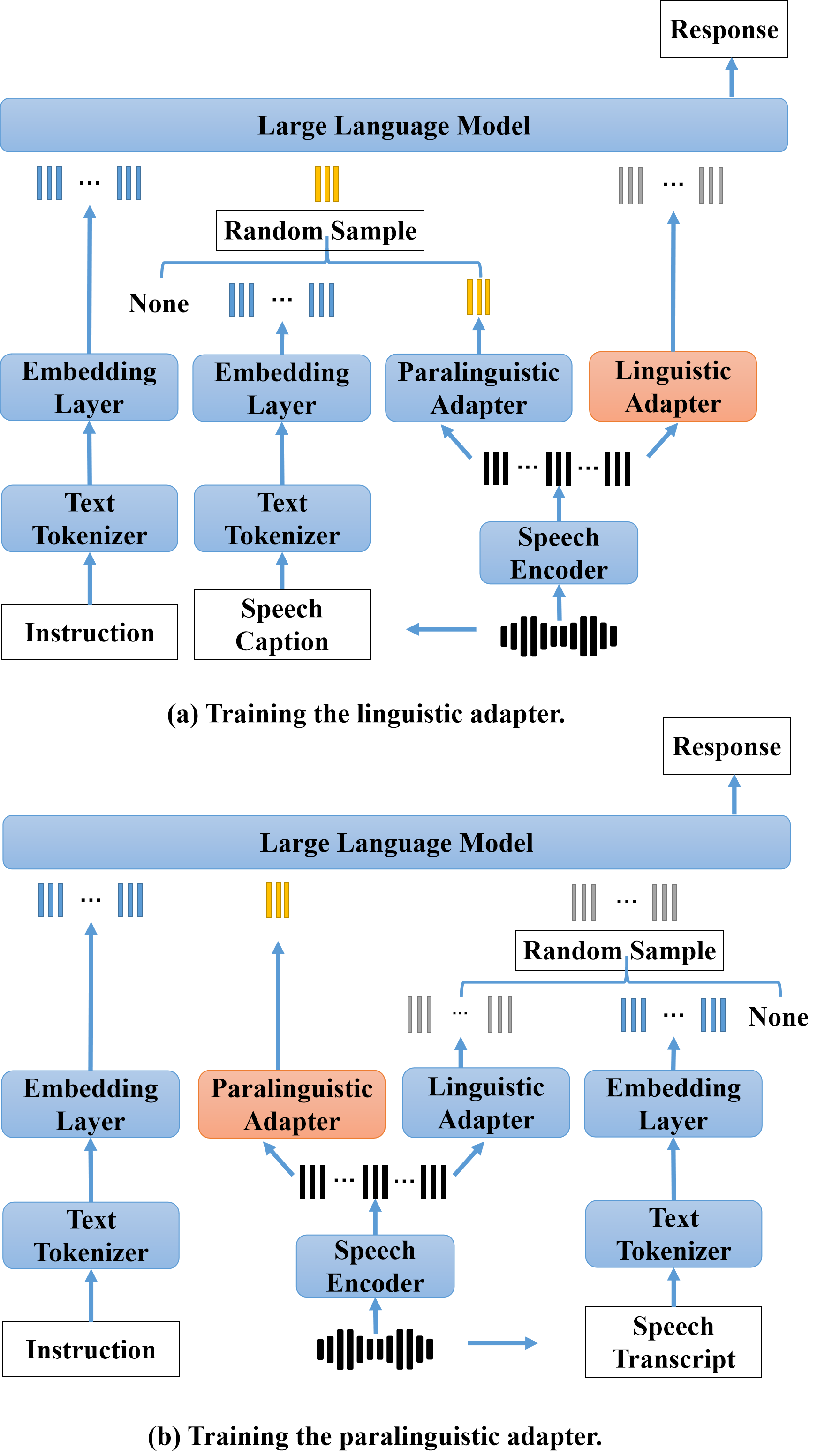}}
\end{minipage}
\caption{Illustration of the Equivalence Replacement Regularization.}
\label{fig:regularization}
\vspace{-1em}
\end{figure}

\textbf{Stage 2}: This stage focuses on achieving disentangled information from the two adapters, ensuring the SLM perceives information from embeddings as intended through the proposed equivalence replacement regularization (ERR). As shown in Fig.~\ref{fig:regularization}(a), when training the linguistic adapter on linguistic tasks, the paralinguistic adapter is kept frozen, and linguistic embeddings are randomly combined with paralinguistic embeddings from speech caption text, speech, or none, with equal probabilities. Since linguistic tasks primarily rely on linguistic information, the source—whether speech caption text, speech, or the absence of paralinguistic embeddings—should not significantly affect the outcome. The same approach is applied to train the paralinguistic adapter, as illustrated in Fig.~\ref{fig:regularization}(b).

Note that the adopted ERR does not strictly train the two adapters independently, acknowledging the challenges of complete disentanglement while allowing flexibility. For instance, the linguistic adapter may encode time-dependent paralinguistic nuances, such as intonation. ERR instead ensures each adapter reliably captures its designated information while minimizing overlap. By making complementary information readily available, each adapter is guided to focus on its target aspect. For example, linguistic content can support emotion recognition. Probabilistic inclusion of linguistic embeddings during paralinguistic adapter training encourages the paralinguistic adapter to prioritize paralinguistic information while avoiding unintended capture of parts of linguistic components. A similar approach is applied to linguistic adapter training.

\textbf{Stage 3}: At this stage, the SLM undergoes holistic fine-tuning, ensuring it can adaptively use both types of information within context, avoiding generating task-specific vectors. This is achieved by training the SLM with dual information tasks, constrained by a portion of paralinguistic and linguistic tasks.

\section{Experimental Setup}
\subsection{Datasets}
Several public datasets are utilized for training. For linguistic tasks, we use the LibriSpeech dataset with its official splits.

For paralinguistic tasks, we primarily use the latest TextrolSpeech dataset \cite{10445879}, with each instance annotated with attributes such as gender, energy, pitch, tempo, and emotion. Due to the limited size of the official test set, we create customized train-validation-test splits from the training set. Further details can be found in Sec.~II of the Appendices. Additionally, the MELD \cite{meld} dataset is included for emotion classification, maintaining its official splits.

For dual information tasks, we use the StyleTalk \cite{lin-etal-2024-advancing} and DailyTalk \cite{10095751} datasets. StyleTalk captures diverse speaking styles, presenting different expressions of the same content, while DailyTalk, adapted from DailyDialog \cite{li-etal-2017-dailydialog}, captures contextual nuances in conversations. Both datasets include multi-turn dialogues. To enhance training, we generate additional samples by truncating conversations at various midpoints and use responses generated by the underlying LLM as targets to maintain natural behavior.

For evaluations, we extend the StyleTalk test split by reversing the roles of the assistant and user. This approach enables a comprehensive assessment, focusing on two key aspects of conversation: generating relevant responses and effectively leading discussions.

For each task, multiple system prompts are prepared. During training and evaluation, a prompt is randomly selected for each speech data instance.

\subsection{Metrics}
To evaluate the perception of SLMs on paralinguistic information, we calculate the Weighted Accuracy (WA) for each attribute classification task. For linguistic information evaluation, we use Word Error Rate (WER) for ASR tasks. 

To evaluate the capabilities of SLMs in emotional conversation, we use two reference-free metrics with an advanced LLM acting as a judge. Following \cite{llamaomni}, we assess SLMs' responses in conversational contexts from two key aspects: content and style, and report CS Scores. The content score measures how well the response addresses the user's utterance, while the style score evaluates how appropriate the response's style is for conversational scenarios. Additionally, following \cite{li2024enhancingemotionalgenerationcapability}, we use the Emotional Generation Score (EGS) to evaluate how well the emotional responses align with human preferences. Based on Goleman’s Emotional Intelligence Theory, the EGS measures four aspects: C1: content relevance; C2: negative emotion avoidance; C3: positive emotion display; C4: positive impact. See Sec.~III in the Appendices for more details.

\subsection{Training Details}
Throughout this paper, we use Whisper-large-v3 \cite{pmlr-v202-radford23a} as the speech encoder, following Qwen2-Audio \cite{Qwen2Audio} and Llama-Omni \cite{llamaomni}. The paralinguistic adapter consists of a single Transformer layer with 8 heads and a dropout rate of 0.1, an adaptive pooling layer, and a linear layer, producing $n_a=10$ paralinguistic embeddings. Following \cite{SLAMASR}, the linguistic adapter has a hidden layer dimension of $d_h=2048$ and a downsampling rate $k=5$, leading to embeddings $\mathbf{E}^{\mathbf{S}}$ at 10Hz.

We trained two SLMs, SLM-Qwen and SLM-Llama, using Qwen2.5-7B-instruction \cite{qwen25} and Llama3.1-8B-Instruction \cite{Llama3}, respectively. All experiments were conducted using an instruction tuning approach with next-token prediction loss. The setup included four A800-80GB GPUs with a batch size of 48. The AdamW optimizer was employed with a maximum learning rate of 5e-5 for stage 1 and 5e-6 for subsequent stages, along with a weight decay of 0.05. A linear decay learning rate scheduler was applied for each stage, with 1000 warmup steps in the first two stages. Each training stage comprised three full epochs, and the best checkpoints were selected based on validation set performance.

\section{Results}

\begin{table}[tb]
\centering
\caption{Results on the TextrolSpeech dataset.}

\begin{tabular}{|cccccc|}
\hline
\multicolumn{1}{|c|}{\textbf{}} &
  \multicolumn{1}{c|}{\textbf{Gender}} &
  \multicolumn{1}{c|}{\textbf{Pitch}} &
  \multicolumn{1}{c|}{\textbf{Tempo}} &
  \multicolumn{1}{c|}{\textbf{Energy}} &
  \textbf{Emotion} \\ \hline
\multicolumn{1}{|c|}{\textbf{}} &
  \multicolumn{1}{c|}{\textbf{WA $\uparrow$}} &
  \multicolumn{1}{c|}{\textbf{WA $\uparrow$}} &
  \multicolumn{1}{c|}{\textbf{WA $\uparrow$}} &
  \multicolumn{1}{c|}{\textbf{WA $\uparrow$}} &
  \textbf{WA $\uparrow$} \\ \hline
\multicolumn{6}{|l|}{\textbf{SLM-Qwen (ours)}} \\ \hline
\multicolumn{1}{|c|}{\textbf{P\&L-Embs}} &
  \multicolumn{1}{c|}{93.16} &
  \multicolumn{1}{c|}{82.17} &
  \multicolumn{1}{c|}{91.24} &
  \multicolumn{1}{c|}{79.34} &
  90.12 \\ \hline
\multicolumn{1}{|c|}{\textbf{P-Embs}} &
  \multicolumn{1}{c|}{\begin{tabular}[c]{@{}c@{}}90.11\\ (-3.05)\end{tabular}} &
  \multicolumn{1}{c|}{\begin{tabular}[c]{@{}c@{}}78.91\\ (-3.26)\end{tabular}} &
  \multicolumn{1}{c|}{\begin{tabular}[c]{@{}c@{}}91.28\\ (+0.04)\end{tabular}} &
  \multicolumn{1}{c|}{\begin{tabular}[c]{@{}c@{}}73.29\\ (-6.05)\end{tabular}} &
  \begin{tabular}[c]{@{}c@{}}90.48\\ (+0.36)\end{tabular} \\ \hline
\multicolumn{6}{|l|}{\textbf{SLM-Llama (ours)}} \\ \hline
\multicolumn{1}{|c|}{\textbf{P\&L-Embs}} &
  \multicolumn{1}{c|}{93.37} &
  \multicolumn{1}{c|}{83.59} &
  \multicolumn{1}{c|}{91.15} &
  \multicolumn{1}{c|}{81.39} &
  90.05 \\ \hline
\multicolumn{1}{|c|}{\textbf{P-Embs}} &
  \multicolumn{1}{c|}{\begin{tabular}[c]{@{}c@{}}91.71\\ (-1.66)\end{tabular}} &
  \multicolumn{1}{c|}{\begin{tabular}[c]{@{}c@{}}79.33\\ (-4.26)\end{tabular}} &
  \multicolumn{1}{c|}{\begin{tabular}[c]{@{}c@{}}91.14\\ (-0.01)\end{tabular}} &
  \multicolumn{1}{c|}{\begin{tabular}[c]{@{}c@{}}72.81\\ (-8.58)\end{tabular}} &
  \begin{tabular}[c]{@{}c@{}}88.81\\ (-1.24)\end{tabular} \\ \hline
\end{tabular}
\label{tab:paralinguistictask}
\vspace{-1em}
\end{table}

\begin{table}[tb]
\centering
\caption{Results on the MELD dataset.}
\begin{tabular}{|c|ccc|c|}
\hline
\textbf{Methods}          & \multicolumn{3}{c|}{\textbf{Learnable Module}}           & \textbf{MELD} \\ \hline
\textbf{} & \multicolumn{1}{c|}{\textbf{Encoder}} & \multicolumn{1}{c|}{\textbf{LLM}} & \textbf{Adapter} & \textbf{WA $\uparrow$} \\ \hline
\textbf{emotion2vec} \cite{emotion2vec}      & \multicolumn{1}{c|}{Y} & \multicolumn{1}{c|}{-} & - & 51.9            \\ \hline
\textbf{Qwen-Audio} \cite{QwenAudio}       & \multicolumn{1}{c|}{Y} & \multicolumn{1}{c|}{N} & Y & 55.7            \\ \hline
\textbf{Qwen2-Audio} \cite{Qwen2Audio}      & \multicolumn{1}{c|}{Y} & \multicolumn{1}{c|}{Y} & Y & 55.3            \\ \hline
\textbf{SLM-Qwen (ours)}  & \multicolumn{1}{c|}{N}  & \multicolumn{1}{c|}{N}  & Y  & 53.3          \\ \hline
\textbf{SLM-Llama (ours)} & \multicolumn{1}{c|}{N}  & \multicolumn{1}{c|}{N}  & Y  & 54.9          \\ \hline
\end{tabular}
\label{tab:MELD}
\vspace{-1em}
\end{table}

\subsection{Paralinguistic and Linguistic Tasks} 
Results in Tab.~\ref{tab:paralinguistictask} show that both of our SLMs effectively perceive paralinguistic information, achieving high performance on five attribute classification tasks. Notably, both SLM-Qwen and SLM-Llama, even when trained only with adapters, perform on par with leading models in emotion recognition on the MELD dataset (see Tab.~\ref{tab:MELD}).

Additionally, results in Tab.~\ref{tab:LibriSpeech} show that both SLM-Qwen and SLM-Llama can effectively perceive linguistic information from speech. For the ASR task, both of our SLMs are trained using only the train-splits of the LibriSpeech dataset with 960 hours, with only adapters being learnable. Despite this, their performance approaches that of ASR-specific SLMs.

Furthermore, the results in Tab.~\ref{tab:paralinguistictask} and Tab.~\ref{tab:LibriSpeech} demonstrate that the proposed ERR effectively guides SLMs to perceive paralinguistic and linguistic information from their respective embeddings as intended. This is evidenced by the observation that on paralinguistic tasks (see Tab.~\ref{tab:paralinguistictask}), both SLMs perform robustly when only paralinguistic embeddings (P-Embs) are present. Meanwhile, on linguistic tasks (see Tab.~\ref{tab:LibriSpeech}), they perform robustly when only linguistic embeddings (L-Embs) are present. It is worth noting that, in most cases, better results are obtained when both paralinguistic and linguistic embeddings (P\&L-Embs) are present. This is anticipated, as linguistic information can aid in solving paralinguistic tasks, and vice versa.

\begin{table}[tb]
  \centering
  \caption{Results on the LibriSpeech dataset.}
  \resizebox{1.0\linewidth}{!}{
      \begin{tabular}{|ccccccc|}
      \hline
      \multicolumn{1}{|c|}{\textbf{Methods}} &
        \multicolumn{3}{c|}{\textbf{Learnable Module}} &
        \multicolumn{1}{c|}{\textbf{ASR}} &
        \multicolumn{2}{c|}{\textbf{WER $\downarrow$}} \\ \hline
      \multicolumn{1}{|c|}{\textbf{}} &
        \multicolumn{1}{c|}{\textbf{Encoder}} &
        \multicolumn{1}{c|}{\textbf{LLM}} &
        \multicolumn{1}{c|}{\textbf{Adapter}} &
        \multicolumn{1}{c|}{\textbf{Hours}} &
        \multicolumn{1}{c|}{\textbf{clean}} &
        \textbf{other} \\ \hline
      \multicolumn{7}{|l|}{\textbf{ASR-Specific SLMs}} \\ \hline
      \multicolumn{1}{|c|}{\textbf{Yu et al.(2024)} \cite{10445874}} &
        \multicolumn{1}{c|}{N} &
        \multicolumn{1}{c|}{N} &
        \multicolumn{1}{c|}{Y} &
        \multicolumn{1}{c|}{960} &
        \multicolumn{1}{c|}{2.3} &
        5.2 \\
      \multicolumn{1}{|c|}{\textbf{SLAM-ASR} \cite{SLAMASR}} &
        \multicolumn{1}{c|}{N} &
        \multicolumn{1}{c|}{N} &
        \multicolumn{1}{c|}{Y} &
        \multicolumn{1}{c|}{960} &
        \multicolumn{1}{c|}{1.9} &
        3.8 \\ \hline
      \multicolumn{7}{|l|}{\textbf{General SLMs}} \\ \hline
      \multicolumn{1}{|c|}{\textbf{SALMONN} \cite{tang2024salmonn}} &
        \multicolumn{1}{c|}{N} &
        \multicolumn{1}{c|}{Y} &
        \multicolumn{1}{c|}{Y} &
        \multicolumn{1}{c|}{1960} &
        \multicolumn{1}{c|}{2.1} &
        4.9 \\
      \multicolumn{1}{|c|}{\textbf{Qwen-Audio} \cite{QwenAudio}} &
        \multicolumn{1}{c|}{Y} &
        \multicolumn{1}{c|}{N} &
        \multicolumn{1}{c|}{Y} &
        \multicolumn{1}{c|}{30K} &
        \multicolumn{1}{c|}{2.0} &
        4.2 \\
      \multicolumn{1}{|c|}{\textbf{Qwen2-Audio} \cite{Qwen2Audio}} &
        \multicolumn{1}{c|}{Y} &
        \multicolumn{1}{c|}{Y} &
        \multicolumn{1}{c|}{Y} &
        \multicolumn{1}{c|}{\textgreater{}30K} &
        \multicolumn{1}{c|}{1.6} &
        3.6 \\ \hline
      \multicolumn{1}{|c|}{\textbf{\begin{tabular}[c]{@{}c@{}}SLM-Qwen(ours)\\ (P\&L-Embs)\\ (L-Embs)\end{tabular}}} &
        \multicolumn{1}{c|}{N} &
        \multicolumn{1}{c|}{N} &
        \multicolumn{1}{c|}{Y} &
        \multicolumn{1}{c|}{960} &
        \multicolumn{1}{c|}{\begin{tabular}[c]{@{}c@{}}2.5\\ 2.5\end{tabular}} &
        \begin{tabular}[c]{@{}c@{}}5.5\\ 5.4\end{tabular} \\ \hline
      \multicolumn{1}{|c|}{\textbf{\begin{tabular}[c]{@{}c@{}}SLM-Llama(ours)\\ (P\&L-Embs)\\ (L-Embs)\end{tabular}}} &
        \multicolumn{1}{c|}{N} &
        \multicolumn{1}{c|}{N} &
        \multicolumn{1}{c|}{Y} &
        \multicolumn{1}{c|}{960} &
        \multicolumn{1}{c|}{\begin{tabular}[c]{@{}c@{}}2.3\\ 2.3\end{tabular}} &
        \begin{tabular}[c]{@{}c@{}}5.1\\ 5.1\end{tabular} \\ \hline
      \end{tabular}
  }
\label{tab:LibriSpeech}
\vspace{-1em}
\end{table}

\begin{table}[t]
\centering
\caption{Results of SLM-Llama on the StyleTalk dataset.}
\begin{tabular}{|c|cccccc|}
\hline
\textbf{Methods} &
  \multicolumn{2}{c|}{\textbf{CS Score}} &
  \multicolumn{4}{c|}{\textbf{EGS Score}} \\ \hline
\textbf{} &
  \multicolumn{1}{c|}{\textbf{Content}} &
  \multicolumn{1}{c|}{\textbf{Style}} &
  \multicolumn{1}{c|}{\textbf{C1}} &
  \multicolumn{1}{c|}{\textbf{C2}} &
  \multicolumn{1}{c|}{\textbf{C3}} &
  \textbf{C4} \\ \hline
 &
  \multicolumn{6}{c|}{\textbf{Judge: Qwen2.5-72B-Instruction}} \\ \hline
\textbf{Qwen2-Audio} \cite{Qwen2Audio} &
  \multicolumn{1}{c|}{3.86} &
  \multicolumn{1}{c|}{3.94} &
  \multicolumn{1}{c|}{8.12} &
  \multicolumn{1}{c|}{9.73} &
  \multicolumn{1}{c|}{7.70} &
  8.16 \\ \hline
\textbf{Llama-Omni} \cite{llamaomni} &
  \multicolumn{1}{c|}{3.76} &
  \multicolumn{1}{c|}{4.00} &
  \multicolumn{1}{c|}{8.10} &
  \multicolumn{1}{c|}{9.64} &
  \multicolumn{1}{c|}{7.55} &
  8.09 \\ \hline
\textbf{SLM-Qwen (ours)} &
  \multicolumn{1}{c|}{\textbf{4.28}} &
  \multicolumn{1}{c|}{\textbf{4.40}} &
  \multicolumn{1}{c|}{\textbf{8.98}} &
  \multicolumn{1}{c|}{\textbf{9.95}} &
  \multicolumn{1}{c|}{\textbf{8.44}} &
  \textbf{8.91} \\ \hline
\textbf{SLM-Llama (ours)} &
  \multicolumn{1}{c|}{4.12} &
  \multicolumn{1}{c|}{4.22} &
  \multicolumn{1}{c|}{8.69} &
  \multicolumn{1}{c|}{9.74} &
  \multicolumn{1}{c|}{8.19} &
  8.60 \\ \hline
 &
  \multicolumn{6}{c|}{\textbf{Judge: Llama-3.1-70B-Instruction}} \\ \hline
\textbf{Qwen2-Audio} \cite{Qwen2Audio} &
  \multicolumn{1}{c|}{3.95} &
  \multicolumn{1}{c|}{4.03} &
  \multicolumn{1}{c|}{8.71} &
  \multicolumn{1}{c|}{9.97} &
  \multicolumn{1}{c|}{7.33} &
  7.97 \\ \hline
\textbf{Llama-Omni} \cite{llamaomni} &
  \multicolumn{1}{c|}{3.80} &
  \multicolumn{1}{c|}{4.06} &
  \multicolumn{1}{c|}{8.51} &
  \multicolumn{1}{c|}{9.91} &
  \multicolumn{1}{c|}{7.04} &
  7.63 \\ \hline
\textbf{SLM-Qwen (ours)} &
  \multicolumn{1}{c|}{\textbf{4.41}} &
  \multicolumn{1}{c|}{\textbf{4.44}} &
  \multicolumn{1}{c|}{\textbf{9.59}} &
  \multicolumn{1}{c|}{\textbf{10.00}} &
  \multicolumn{1}{c|}{\textbf{8.07}} &
  \textbf{8.73} \\ \hline
\textbf{SLM-Llama (ours)} &
  \multicolumn{1}{c|}{4.28} &
  \multicolumn{1}{c|}{4.34} &
  \multicolumn{1}{c|}{9.23} &
  \multicolumn{1}{c|}{9.92} &
  \multicolumn{1}{c|}{7.93} &
  8.46 \\ \hline
\end{tabular}
\label{tab:StyleTalk}
\vspace{-1em}
\end{table}

\subsection{Emotional Conversation}
We compare our SLMs with Qwen2-Audio \cite{Qwen2Audio} and Llama-Omni \cite{llamaomni}, two leading SLMs with conversational context understanding abilities. To avoid bias, we use Qwen2.5-72B-Instruction \cite{qwen25} and Llama-3.1-70B-Instruction \cite{Llama3} as independent judges. As shown in Tab.~\ref{tab:StyleTalk}, both judges consistently give higher scores to our SLMs across all metrics. This demonstrates that SLMs trained with our approach can adaptively use both types of information and understand context. For detailed case comparisons, please refer to Sec.~IV in the Appendices.

\section{Conclusion}
This work presents an efficient approach to extend existing LLMs into SLMs for emotional conversations. By disentangling paralinguistic and linguistic information and avoiding the generation of task-specific vectors, our approach enables SLMs, built from frozen LLMs, to effectively perceive both types of information from speech while maintaining context understanding. This approach only requires training two adapters on common datasets, demonstrating parameter and data efficiency. Our SLMs achieve competitive performance in emotional conversation scenarios, highlighting their ability to adaptively utilize both types of information within contexts.

\bibliographystyle{IEEEbib}
\bibliography{icmeshort}

\end{document}